\RequirePackage{fix-cm}
\documentclass[smallextended]{svjour3}       
\smartqed  
\usepackage{graphicx}
\usepackage[ruled,vlined]{algorithm2e}
\usepackage{amsmath}
\usepackage{amssymb}
\usepackage{multirow}

\usepackage{natbib}
\setcitestyle{authoryear,open={(},close={)}} 

\begin{document}

\title{Exploiting Parallel Corpora to Improve Multilingual Embedding based Document and Sentence Alignment
}


\author{Dilan Sachintha         \and
        Lakmali Piyarathna  \and
        Charith Rajitha \and
        Surangika Ranathunga
}


\institute{Dilan Sachintha \at
              University of Moratuwa, Katubedda 1040, Sri Lanka\\
              \email{dilansachintha.16@cse.mrt.ac.lk}           
           \and
           Lakmali Piyarathna \at
                University of Moratuwa, Katubedda 1040, Sri Lanka\\
              \email{lakmali.16@cse.mrt.ac.lk}
            \and 
            Charith Rajitha  \at 
             University of Moratuwa, Katubedda 1040, Sri Lanka\\
              \email{rajitha.16@cse.mrt.ac.lk}
            \and 
            Surangika Ranathunga \at 
             University of Moratuwa, Katubedda 1040, Sri Lanka\\
              \email{surangika@cse.mrt.ac.lk}  
}

\date{Received: date / Accepted: date}

\maketitle

\begin{abstract}
Multilingual sentence representations pose a great advantage for low-resource languages that do not have enough data to build monolingual models on their own. These multilingual sentence representations have been separately exploited by few research for document and sentence alignment. However, most of the low-resource languages are under-represented in these pre-trained models. Thus, in the context of low-resource languages, these models have to be fine-tuned for the task at hand, using additional data sources. This paper presents a weighting mechanism that makes use of available small-scale parallel corpora to improve the performance of multilingual sentence representations on document and sentence alignment. Experiments are conducted with respect to two low-resource languages, Sinhala and Tamil. Results on a newly created dataset of Sinhala-English, Tamil-English, and Sinhala-Tamil show that this new weighting mechanism significantly improves both document and sentence alignment. This dataset, as well as the source-code, is publicly released.  
\end{abstract}

\section{Introduction}
\label{intro}

Sentence alignment is the process of mapping sentences from a source text to the sentences in the target text, which gives out the same interpretation. Output of sentence alignment is a parallel corpus. This parallel corpus can be between sentences in the same language (e.g.~each sentence pair corresponding to a complex sentence and its simplified version), or sentences in two different languages. In the latter case,  each sentence in the source language is mapped to the corresponding translation of the target language. In this work, bilingual parallel corpus is considered.\\
The process of parallel sentence extraction from large document collections is known as parallel data mining,  parallel text mining, or bitext mining. The most notable uses of parallel sentences is to be used as training data in building multilingual sentence representation~\cite{artetxe2019massively}, and Machine Translation (MT) systems. Performance of MT systems, in particular Neural machine Translation (NMT) systems heavily depends on the availability of a large parallel corpus~\cite{koehn2017six}. Multilingual sentence representations can be used for a multitude of tasks such as cross-lingual document classification and information retrieval~\cite{ruder2019survey}.\\ 
However, most of the sentence alignment techniques expect the availability of a comparable corpus. A comparable corpus is a collection of documents from different languages, which contains potential translations. Document alignment is a possible way to generate comparable corpora. Document alignment can be identified as a task of matching document pairs by computing a score that denotes the likelihood that they are translations of each other ~\cite{buck2016findings}. The content may or may not be sentence aligned. After document alignment, the sentence space to match becomes considerably smaller. Thereafter, sentence alignment can be done on the aligned documents.

The aim of this research is to implement a system that automatically aligns documents from noisy sources, and then use the aligned documents to automatically extract parallel sentences for the language pairs Sinhala-Tamil, Sinhala-English, and Tamil-English. Sinhala and Tamil are considered as low resource languages. They are the official languages in Sri Lanka. Sinhala is limited to the island nation, while Tamil is being used in certain parts of India as well. According to the language categorization by~\cite{joshi2020state}, Sinhala is placed at class 0, and Tamil is placed at class 3. 

Both document and sentence alignment solutions are implemented by exploiting the recently introduced pre-trained multilingual sentence representations (aka multilingual sentence embeddings). Our document alignment system is based on the work of~\cite{el2020massively}. Here, they have introduced a scoring function to calculate the semantic distance between documents by using multilingual sentence embeddings. They have used multiple sentence weighting schemes such as sentence length, inverse document frequency (IDF) and sentence length combined with IDF (SLIDF) to improve the results further. We implemented the same techniques for English-Sinhala, English-Tamil, and Sinhala-Tamil.

Our sentence alignment system is based on the work of~\citet{artetxe2018margin}. They have first obtained sentence embeddings of all target and source side sentences and calculated normalized cosine similarity over nearest neighbours. Calculated score is used as the similarity measurement between a source and a target sentence pair. We implemented the same technique for aforementioned three language pairs.

In that sense, this research can be considered as a case-study that further elaborates the benefit of  multilingual sentence representations for low-resource language processing. Moreover, to the best of our knowledge, this is the first work to exploit multilingual sentence embeddings on a common dataset for both document and sentence alignment. 

Furthermore, we introduce a new weighting mechanism by utilizing existing parallel data sets such as dictionaries, glossaries, and person name bilingual lists to identify the mapping words and sentences in the two languages. From this improved weighting, we were able to gain an average 14\% improvement in recall than the baseline system for document alignment.\\
Similar to document alignment, for sentence alignment also we introduced a weighting scheme using existing parallel data to improve the similarity measurement between a sentence pair. We were able to outperform the baseline system by average 15\% recall for Tamil - English and 1\% for Sinhala - English and 5\% for Sinhala - Tamil. \\
To evaluate the sentence and document alignment systems, a new dataset that consists of document and sentence aligned data for the considered three languages was prepared, along with a hand-aligned golden dataset. This dataset has been publicly released\footnote{https://github.com/kdissa/comparable-corpus} in the hope that it would serve in further research in document and sentence alignment. This is the first manually curated dataset for the considered three languages for document alignment. Compared to the sentence aligned datasets such as those used by~\citet{chaudhary2019low} that are based on subtitles, our corpus is more general-purpose. Moreover, the considered languages belong to three distinct language families (English - Indo European, Tamil - Dravidian, Sinhala - Indo Aryan). Thus this data set is a much tougher benchmark compared to other multilingual datasets such as Text+Berg corpus~\cite{volkchallenges}, and more open-domain compared to data sets such as those based on the Bible.

\section{Related Work}
\label{sec:relatedWork}
The process of extracting a parallel corpus from the web consists of five main tasks; identifying websites having multilingual content, crawling the identified websites, document alignment, sentence alignment, and sentence pair filtering to acquire the desired corpus size or the desired quality.

\subsection{Acquiring Multilingual Content}
Commonly used multilingual content include parliament Hansards in multiple languages, such as the Hansard of the Canadian Parliament~\cite{brown-etal-1991-aligning}, Wikipedia articles~\cite{zweigenbaum-etal-2017-overview,DBLP:journals/corr/abs-1907-05791}, the Bible ~\cite{thompson-koehn-2019-vecalign}, as well as books, novels, and news websites~\cite{Yeong2019}. 

\subsection{Document Alignment}

Automatic document alignment techniques can be categorized as metadata based, translation based, and multilingual sentence embedding based. \\
Early work on document alignment is mostly metadata based~\cite{ resnik2003web,resnik1998parallel, resnik1999mining}. Though these techniques performed well in parallel document extraction, relying on metadata is not possible in many circumstances because the used heuristics are unique to the used dataset and/or resource. In translation based document alignment methods, either the whole document or a part of the target language document is translated into the source language using previously gained linguistic knowledge on source and target languages, and matching documents are identified using characteristics in the resulting documents ~\cite{uszkoreit2010large, Gomes2016}. Hybrid systems improve over these translation based techniques by incorporating heuristics and Information Retrieval techniques~\cite{Dara2016, Mahata2016}. Even though translation based methods were able to score well in document alignment, their performance highly depends on the accuracy of the translation system used.\\
Recently, deep learning models have been explored for the task of document alignment.~\citet{Guo2019} introduced a new neural architecture called Hierarchical Attention Networks (HAN) that captures the insights about the document structure.\\
Very recently,~\citet{el2020massively} proposed a method that utilizes pre-trained multilingual sentence embeddings to calculate the semantic distance between documents in various languages. They have done experiments for high-resource, mid-resource and also low-resource languages. As mentioned earlier, this is the baseline for our research, and more information of this technique can be found in Section \ref{ourapproach}.

\subsection{Sentence Alignment}

Sentence alignment methods proposed so far can be broadly categorized as statistical, Machine Translation based and multilingual sentence embedding based methods.

Early work on sentence alignment is based on sentence statistics such as the correlation of sentence lengths between source and target languages~\cite{brown-etal-1991-aligning, gale-church-1993-program}. However when the correlation between source and target languages decreases, the performance of this approach drops rapidly~\cite{ma-2006-champollion}. Addressing the above-mentioned issue, a hybrid approach is taken, which uses a Machine Translation system and the statistical features of the sentences for alignment~\cite{moore2002fast, varga2007parallel}.

Similar to document alignment, more recent work seems to exploit the pre-trained multilingual sentence representation models.~\citet{artetxe2018margin} use multilingual sentence level embeddings, Neural Machine Translation and supervised classification to identify parallel sentence pairs in French - English corpora. This can be considered as the first research in this line.~\citet{chaudhary2019low} used the LASER toolkit to obtain multilingual sentence representations in their submission to WMT16 shared task. They show that their method outperforms both statistic based and Machine Translation based methods for English - Nepali and English - Sinhala sentence extraction.

In addition to pre-trained multilingual sentence representations, deep learning has also been used for sentence alignment.~\cite{Gregoire2017} developed a deep learning framework for sentence extraction using a Recurrent Neural Network (RNN). This method uses sentence embeddings as input features, and cosine similarity is used to reduce the number of candidates to compare.

\section{Multilingual Sentence Representations}\label{msrepre}

Using word representations or word embeddings is very common in all Natural Language Processing (NLP) tasks today. These word embeddings are able to project syntactic and semantic features of the words into an embeddings space. But when working with sentence level NLP tasks, these word representations are needed to be analyzed at a sentence level. 

According to \citet{schwenk2017learning}, a popular approach to build sentence level representations is to use the ``encoder-decoder approach". Here, the input sentence is encoded to an internal representation and it can be decoded to generate the output sentence. However, all these deep learning techniques are data hungry approaches and therefore a good solution is to first learn language representations on unlabelled data, and then integrating them with task-specific down-stream systems ~\cite{artetxe2019massively}. However, the problem here is that these approaches are specific to the language considered, and are not able to grasp the information across different languages. With this, a new problem arises for low-resource languages because these systems cannot learn the sentence representations due to the lack of data on these languages. Therefore, a system that can produce sentence representations without being limited on the language is highly valuable for low-resource languages.

One such solution is Multilingual-BERT \footnote{https://github.com/google-research/bert/blob/master/multilingual.md}, which supports for around 104 languages. This is the multilingual variant of BERT (Bidirectional Encoder Representations from Transformers), which is a language representation model designed to pre-train deep bidirectional representations from unlabeled text~\cite{devlin2018bert}. Multilingual-BERT model includes Tamil and English, but not Sinhala.

\citet{artetxe2019massively} introduced another solution with a single encoder that can handle multiple languages, so that sentences that are semantically similar lie closer in the embedding space. They used a BiLSTM encoder, which is pre-trained on 93 languages (using parallel corpora), and this encoder is coupled with an auxiliary decoder. Sentence embeddings are obtained by applying a max-pooling operation over the output of the encoder and used to initialize the decoder LSTM through a linear transformation. The encoder and decoder are shared by all the languages and for that, a joint byte-pair encoding (BPE) vocabulary made on the concatenation of all training corpora is used. With this, the encoder is able to learn language independent representations due to the fact that it does not have an input signal about what the language is.

According to \citet{artetxe2019massively}, these generated sentence embeddings can be used on many NLP tasks such as cross-lingual natural language inference, cross-lingual document classification, and bitext mining. They have publicly released their implementation as a toolkit named LASER\footnote{https://github.com/facebookresearch/LASER} with the support for 93 languages among which English, Sinhala and Tamil are included. 

\section{Data set}\label{dataset}
We selected four news web sites that publish the same news in English, Tamil and Sinhala languages as the data source to create the evaluation data set. The selected web sites are Hiru News \footnote{http://www.hirunews.lk}, NewsFirst \footnote{https://www.newsfirst.lk/}, Army News \footnote{https://www.army.lk/} and ITN \footnote{https://www.itnnews.lk}. During pre-processing, news content of each web page was merged into a single string, and image and video tags were removed. To eliminate very short content pages, a threshold of 50 characters was applied.

Some web sites have published each news in all three languages having the same content, sentence structure, order of sentences, and information ﬂow. As an example, for most of the English documents collected from Hiru News and Army News, there were exact translations of the document in Sinhala and Tamil sides. In NewsFirst and ITN, the same news was published in all three languages having the same content but with a relatively low correlation compared to Hiru News. Hence, to balance the data set, we selected a sub set of around 2000 documents based on the published date from the pre-processed documents of each web site to prepare the ground truth. Due to the low correlation between documents published by ITN and Newsfirst they have a lower number of aligned document pairs in the ground truth alignment compared to Army News and Hiru. The number of selected documents from source and target sides and aligned documents pairs in the ground truth alignment for each web site is listed in Table \ref{golden-align-count}. 

In order to create the ground truth alignment, we used some additional information about the data set and the support of human annotators who are ﬂuent in Sinhala, English and Tamil languages. 

\begin{itemize}
  \item URL of each Hiru news document contains a unique id, which is repeated in the URL of the English news article, if the news is published in English language. We used this property to create the ground truth alignment for the Hiru news data set. The created alignment was veriﬁed by one human annotator by going through each aligned document pair in the alignment.
  \item Army news also had the publication date as the shared attribute between the articles of the three languages. The same news was published in all three languages at the exact same date and time. Similar to Hiru news, we created the ground truth alignment for Army news data set using the relationship in publication time, and later veriﬁed the alignment with the help of a human annotator.
  \item Documents crawled from the other two web sites, NewsFirst and ITN did not have any such metadata that we could use to create the ground truth alignment. Therefore, ground truth alignment was manually created by human annotators and was veriﬁed by the same annotators by switching the data sets.
\end{itemize}

Comparable document pairs identified in the ground truth document alignment were used as the input to the sentence alignment system. The number of input sentences in source side and target side for each language pair is listed in Table \ref{Sentence aignment dataset}. Given the large number of sentences in each side, it would take a very long time for human annotators to find all sentence pairs that are translations of each other. Therefore we created a ground truth alignment for sentence alignment data set including roughly 300 one-to-one sentence pairs from each web site in all three language pairs except ITN Tamil-English, ITN Sinhala-Tamil and Newsfirst Sinhala-Tamil. As there was a small number of aligned document pairs for those two websites for the above mentioned language pairs, we cloud not find 300 aligned sentence pairs. The number of aligned sentence pairs in the ground truth alignment for each web site is listed in Table~\ref{tab:golden-sentence-alignments}.

To evaluate the impact of sentence alignment system on downstream MT quality, a hold-out test set of Wikipedia translations for Sinhala – English from the FLORES dataset~\cite{guzman-etal-2019-flores}, and for Tamil-English and Sinhala - Tamil a hold-out test set from the parallel data set created by the National Languages Processing Center of University of Moratuwa, Sri Lanka\footnote{https://uom.lk/nlp}  are used.

\begin{table}
  \caption{Document alignment data set with golden alignment counts}
  \label{golden-align-count}
  \begin{tabular}{|c|c|c|c|c|c|c|c|c|c|}
    \hline
    Website&\multicolumn{3}{c}{SI - EN}&\multicolumn{3}{c}{TA - EN}&\multicolumn{3}{c}{SI - TA}\\
   \hline
    &SI&EN&Aligned&TA&EN&Aligned&SI&TA&Aligned \\
    \hline
    Army &  2033 & 2081 & 1848 & 1905 & 2081 & 1671 & 2033 & 1905 & 1578\\
     \hline
    Hiru &  3133 & 1634 & 1397 & 2886 & 1634 & 1056 & 3133 & 2886 & 2002 \\
    \hline
    ITN & 4898 & 1942 & 352 & 1521 & 1942 & 112 & 4898 & 1521 & 34\\
   \hline
    Newsfirst & 1819 & 2278 & 344 & 2333 & 2278 & 316 & 1819 & 2333 & 97\\
  \hline
\end{tabular}
\end{table}

\begin{table}
  \caption{Sentence alignment data set}
  \label{Sentence aignment dataset}
  \begin{tabular}{|c|c|c|}
     \hline
    Language pair&Source sentences&Target sentences \\
     \hline
    SI - EN & 153750 & 140701 \\
     \hline
    TA - EN & 87266 & 87330 \\
     \hline
    SI - TA & 38101 & 37371 \\
  \hline
\end{tabular}
\end{table}

\begin{table}
  \caption{Ground truth alignment counts - Sentence Alignment}
  \label{tab:golden-sentence-alignments}
  \begin{tabular}{|c|c|c|c|}
    \hline
    Web site & SI - EN & TA - EN & SI - TA \\
    \hline
    Army news & 300 & 300 & 300 \\
    \hline
    ITN & 300 & 287 & 78 \\
    \hline
    Hiru & 300 & 300 & 300 \\
    \hline
    Newsfirst & 300 & 300 & 169 \\
    \hline
\end{tabular}
\end{table}

For our improvements, we used already existing parallel data sets such as dictionaries, glossaries, and person name bilingual lists, for the considered languages. These data sets were taken from Natural Language Processing Centre of University of Moratuwa~\cite{8421901}. Samples of these data sets are shown in Table~\ref{tab:dictionary-overview}.

\textbf{Person Names Bilingual List}
This dataset contains person names in all three languages. This dataset contains 3 sets of two way parallel data. The English-Sinhala dataset has 6194 parallel names, English-Tamil contains 1374 parallel names and Sinhala-Tamil dataset contains 76334 parallel names.

\textbf{Designations Bilingual List}
This also has designations in all three languages in 3 sets. The English-Sinhala dataset has 6764 parallel designations, English-Tamil dataset has 5779 parallel designations and the Sinhala-Tamil dataset contains 44193 parallel designations.

\textbf{Word Dictionary}
This word dictionary also contains words and their translation in the three languages as three sets. The English-Sinhala dictionary has 23722 parallel words, English-Tamil dictionary has 36551 parallel words and the Sinhala-Tamil dictionary has 19132 words.

\textbf{Glossary}
This glossary is three way parallel unlike the other data sets that we got our hands on. This glossary contains short phrases rather than words. It contains 24261 phrases in all three languages.

\begin{table}
  \caption{Overview of the Dictionaries}
  \label{tab:dictionary-overview}
  \centering
  \includegraphics[width = 1.0\textwidth]{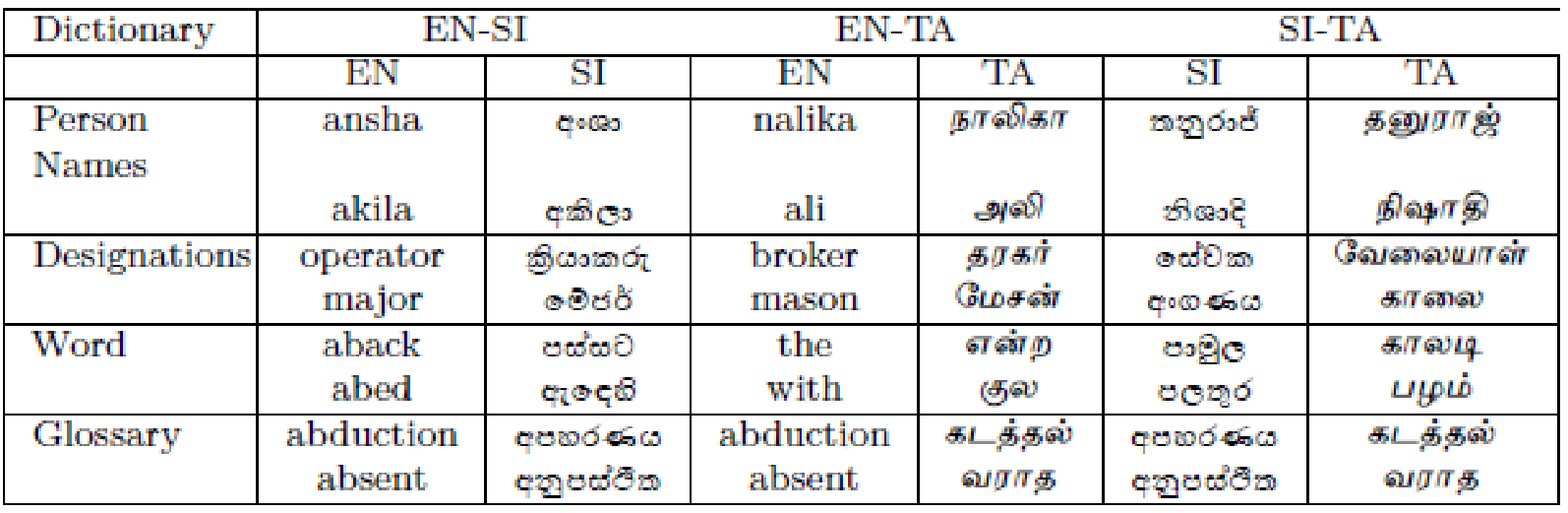}
\end{table}  
\section{Methodology}\label{ourapproach}
\subsection{Document Alignment}

For document alignment,~\citet{el2020massively} have done an impressive work on both high resource languages and low resource languages. Here, they have experimented on document alignment for English-Tamil documents as well (however, the corpus they used is not publicly available). We recreated this solution for parallel document extraction for English-Sinhala, English-Tamil and Sinhala-Tamil.

\subsubsection{Baseline Implementation}
This implementation introduces an unsupervised scoring function that uses multilingual sentence embeddings to calculate the semantic distance between two documents. LASER toolkit is used when generating the multilingual sentence embeddings. This calculated distance is used in the below described alignment algorithm that extracts the best matching document pairs.

To calculate the semantic distance, a novel distance metric named Cross-Lingual Sentence Mover's Distance (XLSMD) has been introduced. XLSMD is a distance metric based on Earth Mover's Distance (EMD). XLSMD represents each document as a bag-of-sentences (BOS) with all the sentences containing a pre-calculated probability mass (weight). Equation \ref{eq1} shows the semantic distance between documents $A$ and $B$.
\begin{equation} \label{eq1}
XLSMD(A, B) = \min_{T \geqslant 0} \sum_{i=1}^{V} \sum_{j=1}^{V} T_{i, j} \times \Delta(i, j)
\end{equation}
\begin{equation}
Here, \forall i \sum_{j=1}^{V} T_{i, j} = d_{A, i}   \;\;,\;\;  \forall j \sum_{i=1}^{V} T_{i, j} = d_{B, j}
\end{equation}

Here, $\Delta(i, j)$ is the Euclidean distance between the two sentence embeddings. In Equation \ref{eq1}, $T_{i,j}$ shows how much of sentence $i$ in document $A$ is assigned to sentence $j$ in document $B$ (probability mass of a sentence).
\begin{equation}\label{eq2}
    d_{A,i} = \dfrac{count(i)}{\sum_{s\in A}count(s)}
\end{equation}
Equation \ref{eq2} shows the first function used for the probability mass. Here, they have used the relative frequencies of sentences as the probability mass. In this, $\sum_{s\in A}count(s)$ represents the sentence count in document $A$. After calculating XLSMD, the distance is used in the alignment algorithm.

To make the XLSMD calculations more tractable, a greedy algorithm named Greedy Mover's Distance is introduced. Here, the algorithm first calculates the Euclidean distance between each sentence pair and sorts them in ascending order. Then it iteratively multiplies each distance by the smallest weight among the two sentences, which is named as the \emph{flow} value as shown in Equation \ref{eq3}.
\begin{equation}\label{eq3}
    distance = distance + ||s_{A} - s_{B}|| \times flow \times w_{A,B,i}
\end{equation}
Later, they introduced the following advanced weighting schemes in-place of relative frequency.\\

\textbf{Sentence Length Weighting}

This weighting scheme is used under the assumption that longer sentences should be given more probability mass than shorter sentences.
Equation \ref{eq4} defines how this weight is calculated.
\begin{equation}\label{eq4}
    d_{A,i} = \dfrac{count(i)\times|i|}{\sum_{s\in A}count(s)\times|s|}
\end{equation}
Here, $|i|$ and $|s|$ represent the number of tokens in the sentences $i$ and $s$, respectively.\\

\textbf{IDF Weighting}

IDF stands for Inverse Document Frequency. Here, they have used the argument that the sentences that occur more frequently in the corpus should be given less importance than the rare sentences in the document.
Equation \ref{eq5} defines how it is calculated.
\begin{equation}\label{eq5}
    d_{A,i} = 1 + \log\dfrac{N + 1}{1 + |{d\in D:s\in d}|}
\end{equation}
Here, $N$ is the total number of documents in domain $D$, and $|{d\in D:s\in d}|$ is the number of sentence that contain sentence $s$.\\

\textbf{SLIDF Weighting}

In this scheme, both the above schemes are joined together to form a joint weighting scheme.
\subsubsection{Our Improvements}
In our implementation, we did several improvements to the above baseline by considering the nature of data, and by exploiting the available parallel language resources.

\textbf{Datewise Filtering}

Since our dataset is completely taken from news websites, all the news documents have the published date as a metadata. Moreover, in most of the cases the same news document is published in all three languages in the same day. Therefore, before starting the aligning process, we filtered and divided the documents using the published date and reduced the search space by a considerable amount.

\textbf{Exploiting the Available Parallel Dictionaries}

The three weighting schemes used by ~\citet{El-Kishky2020} are generated from the dataset used for alignment. We used already available parallel datasets mentioned in Section \ref{dataset} to introduce an additional weighting scheme on top of the above schemes. Here, if a sentence $s_{A}$ from document $A$ contains a word $w$ in the parallel dataset and the sentence $s_{B}$ from document $B$ contains the translation of the word $w$, a counter value is incremented and inserted into the Greedy Mover's Algorithm. The inserted value is calculated by Equation \ref{eq7}.
\begin{equation}\label{eq7}
    w_{A,B,i} = \dfrac{|s_{A}| - count}{|s_{A}|} \;\;\;\;\;\;\;\;\;|s_{A}| = Number ~of ~tokens ~in ~sentence ~s_{A}
\end{equation}
This $W_{A,B,i}$ is inserted into the Greedy Mover's Distance algorithm as shown in Equation \ref{eq8}.
\begin{equation}\label{eq8}
    distance = distance + ||s_{A} - s_{B}|| \times flow \times w_{A,B,i}
\end{equation}
This way, when more words that map with the parallel data sets are identified in a sentence pair, the distance between the two sentences will be lesser.

\paragraph{Usage of Person Names Bilingual List\newline}
 We added the parallel words in person names bilingual list into a dictionary structure where keys are words from language $A$ and the values are arrays of translations of the key in language $B$ (One person name has multiple translations sometimes due to multiple types of spelling formats). When calculating the weights, for each sentence pair, we iterated through the words in the sentence to calculate the mapping counts. Here, we split the sentence $s_{A}$ into words and check if each word $w$ exists in the dictionary. If it exists, we get the parallel words $v_{B}$, and check if each parallel word exists in the sentence $s_{B}$. If so, we increase the counter and remove the mapped word from the sentence $s_{B}$. This counter value is used as the input in Equation \ref{eq7}. Algorithm \ref{algPersonNames} explains this process.

\paragraph{Usage of Designations Bilingual List and Word Dictionary\newline}
The difference in these datasets from the above dataset is that designations and word dictionary sometimes contain more than one word (contains phrases). Therefore, when calculating weights, we implemented a separate algorithm that identifies the multiple word mappings considering the multiple words. Here, for each sentence $s_{A}$, we get all the permutations of words from length one to length five (maximum length of a record in the dictionary is five). Then we do the same process described above to get the mapping counts.  Algorithm \ref{algDesigWord} depicts this process. When person names, designations, and word dictionary are used in combination, we sum up the counter values from both Algorithm \ref{algPersonNames} and \ref{algDesigWord}, and use that value as the input for Equation~\ref{eq7}.

\begin{minipage}{5.4cm}
\begin{algorithm}[H]
\SetAlgoLined
\SetKwInOut{Input}{Input}
\SetKwInOut{Output}{Output}
\Input{$s_{A}$, $s_{B}$, $dict$}
\Output{$w_{A,B,i}$}
 $w_{A}, w_{B} \gets list(s_{A}), list(s_{B})$\;
 $count \gets 1$\;
 \For{$w \in w_{A} : |w| = 1$}{
  \If{$w \in dict$}{
    $v_{B} \gets dict[w]$\;
    \For{$v \in v_{B}$}{
        \If{$v \in w_{B}$}{
            $count \gets count + 1$\;
            Remove $w$ from $w_{B}$\;
        }
    }
  }
 }
 \caption{Algorithm for checking Person names}
 \label{algPersonNames}
\end{algorithm}
\end{minipage}
\hfill
\begin{minipage}{5.8cm}
 \begin{algorithm}[H]
\SetAlgoLined
\SetKwInOut{Input}{Input}
\SetKwInOut{Output}{Output}
\Input{$s_{A}$, $s_{B}$, $dict$}
\Output{$w_{A,B,i}$}
 $w_{A}, w_{B} \gets list(s_{A}), list(s_{B})$\;
 $count \gets 1$\;
 \eIf{$|w_{A}| \geqslant 5$}{
    \For{$w \in w_{A}: |w| = 1, 2, 3, 4, 5$}{
        \If{$w \in dict$}{
            $v_{B} \gets dict[w]$\;
            \For{$v \in v_{B}$}{
                \If{$v \in w_{B}$}{
                    $count \gets count + 1$\;
                    Remove $w$ from $w_{B}$\;
                }
            }
        }
    }
 }{
    Algorithm \ref{algPersonNames}
 }
 \caption{Algorithm for checking Designations and Word Dictionary}
 \label{algDesigWord}
\end{algorithm}
\end{minipage}

\paragraph{Improved Dictionary\newline}
We also experimented by using the glossary mentioned in Section \ref{dataset} and there was no any considerable amount of improvement to be seen. This was due to the fact that the glossary mostly contained sentences and the respective translations rather than words. Therefore, when the glossary is cross-checked with the sentence pairs, the number of mappings is very low. Therefore we utilized the parallel sentences in the glossary to generate new parallel words.  We used the word dictionary that we had, and cross-checked the sentences in the glossary with the words in the dictionary. We removed the parallel words that we found in the glossary sentences and extracted the remaining words from both languages as a parallel record. This way, the amount of words in one record in the glossary got reduced in a considerable amount and we were able to improve the existing dictionary by adding the records we found from the glossary to the word dictionary. An overview of the improved dictionary is shown in Table \ref{tab:improveddic}.

\begin{table}
  \caption{Overview of the Improved Dictionary}
  \label{tab:improveddic}
  \centering
  \includegraphics[width = 1.0\textwidth]{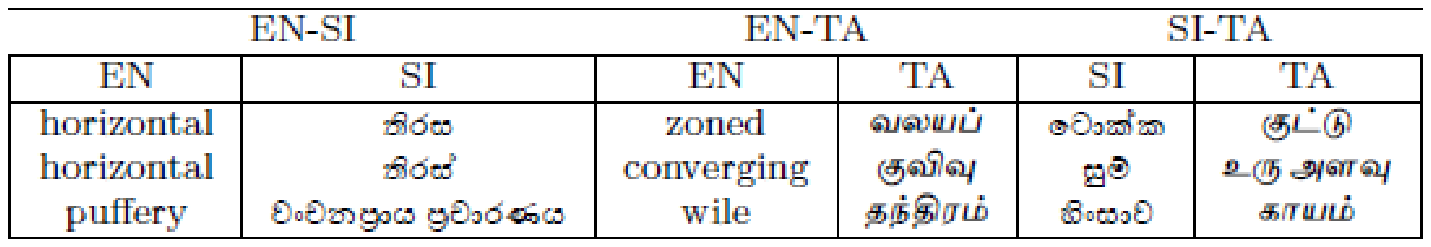}
\end{table}  
\subsection{Sentence Alignment}
\subsubsection{Baseline system}
Our baseline system is based on the sentence alignment system proposed by ~\citet{artetxe2018margin}. They have first learned multilingual sentence embeddings using an encoder-decoder architecture and existing parallel data. In here also, we used the publicly available LASER toolkit. Sentences in both source and target sides were converted into sentence embeddings, and each sentence in source side is scored with all the sentences in the target side. Cosine similarity of the two vector representations is considered as the similarity score of the respective two sentences.

Below three strategies proposed by~\citet{artetxe2018margin} were used for candidate generation. 

\begin{itemize}
  \item Forward : Each source sentence is aligned with exactly one best scoring target sentence. Some target sentences may be aligned with multiple source sentences or with none.
  \item Backward : Equivalent to the forward strategy, but going in the opposite direction.
  \item Intersection : Intersection of forward and backward candidates, which discards sentences with inconsistent alignments.
\end{itemize}

These candidates are then sorted according to their scores, and a threshold is applied. This can be either optimized on the development data, or adjusted to obtain the desired corpus size.

\subsubsection{Our Improvements}
In our implementation, we improved the similarity measurement by introducing a weighting scheme using the existing parallel data described in Section \ref{dataset}. We combined the improved dictionary together with the person names bilingual list and designations bilingual list as the parallel dataset as it gave the best results for document alignment.
In forward candidate generation strategy, we used cosine similarity as the initial similarity score and selected the best matching k \footnote{We use k=4 for all experiments in this work as it gave the best results in all our experiments.} candidates over cosine similarity for each source sentence. Here, if the source sentence $s_{A}$ from document $A$ contains a word $w$ in the parallel dataset and the target sentence $s_{B}$ from the selected k candidates contains the translation of the word $w$, a counter value is incremented. This counter value is used to calculate the weight using Equation \ref{eq13} (Multiplicative inverse of Equation \ref{eq7}), which gives a higher weight for sentence pairs having more overlapping tokens and a lower weight for sentence pairs with a lower number of overlapping tokens.
\begin{equation}\label{eq13}
    w_{A,B} = \dfrac{|s_{A}|}{|s_{A}| - count} \;\;\;\;\;\;|s_{A}| = Number ~of ~tokens ~in ~ source ~sentence ~s_{A}
\end{equation}

New similarity score between each source sentence $s_{A}$ and each target sentence $s_{B}$ in its $k$ selected candidates is calculated using Equation \ref{eq12},
\begin{equation}\label{eq12}
    similarity\_score_{a,b} = cosine\_similarity_{a,b} \times w_{a,b}
\end{equation}

Then each source sentence is aligned with the best scoring target sentence according to the above calculated similarity scores.

\section{Evaluation}\label{evaluation}
We evaluated our implementations separately for document alignment and sentence alignment.

\subsection{Document Alignment}
For document alignment, we used the same method that the baseline was evaluated on \cite{el2020massively}. Here, only recall (i.e. what percentage of the aligned document pairs in the golden alignment set is found by the system) is used to evaluate the generated pairs because the dataset contains many more parallel document pairs that are not in the ground truth set.

We did experiments for English-Sinhala, English-Tamil and Sinhala-Tamil with selected news web sites. The results are shown in Table \ref{tab:resultsdocalign}.

\begin{table}
  \caption{Document Alignment Results. SL - Sentence Length, IDF - Inverse Document Frequency, SLIDF - SL + IDF, NF - Newsfirst, A - Baseline system, B - Using person names bilingual list, C - Using designations bilingual list on top of B, D - Using word dictionary on top of C, E - Using improved dictionary on top of C. Baseline and other results are taken after applying date-wise filtering.}
  \label{tab:resultsdocalign}
  \begin{tabular}{|c|c|c|c|c|c|c|c|c|c|c|c|c|}
     \hline
    &\multicolumn{4}{c}{English-Sinhala}&\multicolumn{4}{c}{English-Tamil}&\multicolumn{4}{c}{Sinhala-Tamil}\\
     \hline
    &Hiru&ITN&NF&Army&Hiru&ITN&NF&Army&Hiru&ITN&NF&Army\\
     \hline
    SL&0.82&0.85&0.88&0.99&0.26&0.44&0.41&0.69&0.45&0.41&0.63&0.83\\
    IDF&0.78&0.84&0.81&0.98&0.24&0.44&0.37&0.57&0.42&0.47&0.59&0.73\\
    SLIDF&0.82&0.85&0.88&0.99&0.26&0.44&0.42&0.69&0.46&0.44&0.63&0.84\\
     \hline
    SL&0.85&0.87&0.88&0.99&0.28&0.43&0.43&0.71&0.50&0.50&0.70&0.86\\
    IDF&0.81&0.84&0.79&0.98&0.25&0.39&0.35&0.57&0.45&0.50&0.62&0.73\\
    SLIDF&0.85&0.87&0.86&0.99&0.28&0.43&0.43&0.71&0.50&0.50&0.70&0.86\\
     \hline
    SL&0.85&0.87&0.88&0.99&0.31&0.47&0.46&0.71&0.50&0.50&0.71&0.86\\
    IDF&0.82&0.85&0.80&0.98&0.29&0.41&0.47&0.58&0.46&0.50&0.62&0.73\\
    SLIDF&0.86&0.87&0.87&0.99&0.31&0.47&0.46&0.71&0.50&0.50&0.71&0.86\\
     \hline
    SL&0.85&\textbf{0.87}&0.88&0.99&0.55&0.68&0.67&0.83&0.53&0.50&0.73&0.87\\
    IDF&0.83&0.86&0.81&0.98&0.46&0.66&0.54&0.71&0.48&0.58&0.64&0.75\\
    SLIDF&0.86&\textbf{0.87}&0.88&0.99&0.51&0.68&0.68&0.83&0.53&0.50&0.73&0.87\\
     \hline
    SL&0.85&0.85&0.88&0.99&\textbf{0.60}&\textbf{0.70}&\textbf{0.71}&0.84&\textbf{0.56}&0.61&\textbf{0.78}&0.89\\
    IDF&0.83&0.86&0.83&0.98&0.48&0.67&0.58&0.72&0.52&\textbf{0.64}&0.70&0.77\\
    SLIDF&\textbf{0.86}&0.85&\textbf{0.90}&0.99&0.55&\textbf{0.70}&\textbf{0.71}&\textbf{0.85}&\textbf{0.56}&0.61&\textbf{0.78}&\textbf{0.90}\\
     \hline
\end{tabular}
\end{table}

We found out that in most of the cases, the use of dictionaries outperform the baselines in all the three language pairs. For English-Sinhala, using the improved dictionary combined with person names and designations score well in Hiru and Newsfirst. For Army, we cannot see an improvement because it already has a score of 99\%. But for ITN, a difference can be seen where using improved dictionary does not perform than the existing dictionary.

For both English-Tamil and Sinhala-Tamil, we can see a considerable amount of improvement in using all the dictionaries. Unlike English-Sinhala, both the baselines perform very low in these 2 pairs. The reason for this must be due to the low training data used for Tamil language when training the LASER toolkit ~\cite{artetxe2019massively}, and the highly agglutinating nature of the language. Therefore, by using weights based on dictionaries, we were able to get an improvement of about 20\%.

Even though the person names bilingual list of Sinhala-Tamil has a content about 10 times the other person names bilingual lists, we cannot see a considerable improvement in Sinhala-Tamil compared to other two. This is because most of the common person names already exist in all three parallel sets and not much difference is made even though the rare names exist.

\subsection{Sentence Alignment}
We used two techniques to evaluate the sentence alignment system. First we evaluated the obtained results against a ground truth data set. As this ground truth alignment only contained a small fraction of parallel sentences, there can be many more valid cross lingual sentence pairs in these data sets. Therefore we evaluated aligned sentence pairs using recall (i.e. what percentage of the sentence pairs in the golden alignment set are found by the algorithm), which is one of the commonly used measurements for sentence alignment accuracy ~\cite{thompson-koehn-2019-vecalign, artetxe2018margin}.

Secondly, we used the same evaluation technique used in the WMT’19 sentence pair filtering task ~\cite{inproceedings} to measure improvements in downstream Machine Translation quality. Candidate sentences are scored using a margin based score over cosine similarity, and these scores are used to sub-sample sentence pairs that amount to 1 million target side words. The quality of the resulting subsets is determined by translating the evaluation datasets mentioned in Section~\ref{dataset} using the NMT system fairseq ~\cite{ott-etal-2019-fairseq} trained on this data. The quality of the Machine Translated text is measured by BLEU score using SacreBLEU.

\begin{table}
  \caption{Sentence Alignment Results - Recall values with golden alignment. F - Forward, B - Backward, I - Intersection, BL - Baseline, Dict - Using improved dictionary together with person names bilingual list and designations bilingual list.}
  \label{tab:results-sentence-alignment}
  \begin{tabular}{|c|c|c|c|c|c|c|c|c|c|c|c|c|c|}
    \hline
    {Lang pair}&&\multicolumn{3}{c}{Army News}&\multicolumn{3}{c}{Hiru}&\multicolumn{3}{c}{ITN}&\multicolumn{3}{c}{Newsfirst}\\
    \hline
    &&F&B&I&F&B&I&F&B&I&F&B&I\\
    \hline
    {SI-EN}&BL&0.87&\textbf{0.90}&0.83&\textbf{0.94}&\textbf{0.91}&0.88&0.89&0.92&0.85&0.93&\textbf{0.92}&0.88\\
    &Dict&\textbf{0.89}&\textbf{0.90}&\textbf{0.84}&\textbf{0.94}&0.90&\textbf{0.88}&\textbf{0.93}&\textbf{0.93}&\textbf{0.88}&\textbf{0.94}&\textbf{0.92}&\textbf{0.90}\\
    \hline
    {TA-EN}&BL&0.44&0.61&0.37&0.40&0.53&0.28&0.33&0.59&0.28&0.38&0.60&0.30\\
    &Dict&\textbf{0.60}&\textbf{0.69}&\textbf{0.53}&\textbf{0.58}&\textbf{0.57}&\textbf{0.40}&\textbf{0.51}&\textbf{0.69}&\textbf{0.43}&\textbf{0.61}&\textbf{0.74}&\textbf{0.54}\\
    \hline
    {SI-TA}&BL&0.65&0.74&0.58&0.70&0.74&0.59&0.63&0.63&0.44&0.71&0.75&0.62\\
    &Dict&\textbf{0.67}&\textbf{0.75}&\textbf{0.59}&\textbf{0.75}&\textbf{0.81}&\textbf{0.66}&\textbf{0.69}&\textbf{0.69}&\textbf{0.50}&\textbf{0.76}&\textbf{0.83}&\textbf{0.66}\\
  \hline
\end{tabular}
\end{table}

\begin{table}
  \caption{Sentence Alignment Results - BLEU scores}
  \label{tab:results-sentence-alignment-blue}
  \begin{tabular}{|c|c|c|c|c|c|c|c|c|c|}
    \hline
    {Weighting scheme}&\multicolumn{3}{c}{SI - EN}&\multicolumn{3}{c}{TA - EN}&\multicolumn{3}{c}{SI - TA}\\
   \hline
    &F&B&I&F&B&I&F&B&I\\
     \hline
    Baseline&7.54&\textbf{7.92}&7.67&6.33&\textbf{7.06}&7.17&\textbf{0.41}&0.86&1.29\\
     \hline
    Dictionary weighting&\textbf{7.66}&\textbf{7.92}&\textbf{7.87}&\textbf{6.44}&6.49&\textbf{7.27}&0.33&\textbf{1.2}&\textbf{2.18}\\
   \hline
\end{tabular}
\end{table}

Our sentence alignment system outperforms the baseline system in all three language pairs for all the web sites with the exception of very few as seen in Tables~\ref{tab:results-sentence-alignment} and~\ref{tab:results-sentence-alignment-blue}. Tamil - English language pair shows the highest improvement by outperforming the baseline system by on average 15\% recall. Sinhala - Tamil and Sinhala - English language pairs show on average 5\% and 1\% recall gains respectively. 

Baseline sentence alignment results for both Tamil - English and Sinhala - Tamil language pairs are considerably low compared to Sinhala - English. The low amount of training data used for Sinhala and Tamil when training the LASER toolkit could be the reason for that ~\cite{artetxe2019massively}. Language diversity, and the different forms of inflectional natures of Sinhala and Tamil also contribute to this problem.

All the BLEU scores reported for Sinhala -Tamil are in the range of 0.33 to 2.18, while other two language pairs have 6.5+ BLEU points for all candidate generation strategies. Compared to other two language pairs, input data set for Sinhala - Tamil had a very small amount of sentences in both source and target sides, causing the huge loss in the final results. 

\section{Conclusion}\label{conclusion}
This research presented a unified document and sentence alignment system for the low resource language pairs Sinhala-English, Sinhala-Tamil, and Tamil-English using multilingual sentence representations. To the best of our knowledge, this is the first system that implements both types of alignment on a common dataset using multilingual sentence representations. Furthermore, we were able to exploit the already existing parallel data to fine-tune the alignment done using the pre-trained multilingual sentence representations, which overcomes (at least to some extent) the inability of the same to be easily retrained with new data. In addition, we have publicly released a multilingual dataset that can be used for further research on document and sentence alignment.

For our future work, we are going to try and integrate different distance metrics like cosine similarity along with the dictionary improvements and explore more on different weighting and normalizing schemes. And we are going to see whether the novel idea of integrating parallel lists to improve document and sentence alignment by embeddings can be generalized to other languages.


%
%

\bibliography{template.bib}   

\begin{thebibliography}{34}
\providecommand{\natexlab}[1]{#1}
\providecommand{\url}[1]{\texttt{#1}}
\expandafter\ifx\csname urlstyle\endcsname\relax
  \providecommand{\doi}[1]{doi: #1}\else
  \providecommand{\doi}{doi: \begingroup \urlstyle{rm}\Url}\fi

\bibitem[Artetxe and Schwenk(2018)]{artetxe2018margin}
M.~Artetxe and H.~Schwenk.
\newblock Margin-based parallel corpus mining with multilingual sentence
  embeddings.
\newblock \emph{arXiv preprint arXiv:1811.01136}, 2018.

\bibitem[Artetxe and Schwenk(2019)]{artetxe2019massively}
M.~Artetxe and H.~Schwenk.
\newblock Massively multilingual sentence embeddings for zero-shot
  cross-lingual transfer and beyond.
\newblock \emph{Transactions of the Association for Computational Linguistics},
  7:\penalty0 597--610, 2019.

\bibitem[Brown et~al.(1991)Brown, Lai, and Mercer]{brown-etal-1991-aligning}
P.~F. Brown, J.~C. Lai, and R.~L. Mercer.
\newblock Aligning sentences in parallel corpora.
\newblock In \emph{29th Annual Meeting of the Association for Computational
  Linguistics}, pages 169--176, Berkeley, California, USA, June 1991.
  Association for Computational Linguistics.
\newblock \doi{10.3115/981344.981366}.
\newblock URL \url{https://www.aclweb.org/anthology/P91-1022}.

\bibitem[Buck and Koehn(2016)]{buck2016findings}
C.~Buck and P.~Koehn.
\newblock Findings of the wmt 2016 bilingual document alignment shared task.
\newblock In \emph{Proceedings of the First Conference on Machine Translation:
  Volume 2, Shared Task Papers}, pages 554--563, 2016.

\bibitem[Chaudhary et~al.(2019)Chaudhary, Tang, Guzm{\'a}n, Schwenk, and
  Koehn]{chaudhary2019low}
V.~Chaudhary, Y.~Tang, F.~Guzm{\'a}n, H.~Schwenk, and P.~Koehn.
\newblock Low-resource corpus filtering using multilingual sentence embeddings.
\newblock \emph{arXiv preprint arXiv:1906.08885}, 2019.

\bibitem[Dara and Lin(2016)]{Dara2016}
A.~A. Dara and Y.-C. Lin.
\newblock {YODA System for WMT16 Shared Task: Bilingual Document Alignment}.
\newblock In \emph{Proceedings of the First Conference on Machine Translation:
  Volume 2, Shared Task Papers}, volume~2, pages 679--684, Stroudsburg, PA,
  USA, 2016. Association for Computational Linguistics.
\newblock \doi{10.18653/v1/W16-2366}.
\newblock URL \url{http://aclweb.org/anthology/W16-2366}.

\bibitem[Devlin et~al.(2018)Devlin, Chang, Lee, and Toutanova]{devlin2018bert}
J.~Devlin, M.-W. Chang, K.~Lee, and K.~Toutanova.
\newblock Bert: Pre-training of deep bidirectional transformers for language
  understanding.
\newblock \emph{arXiv preprint arXiv:1810.04805}, 2018.

\bibitem[El-Kishky and Guzm{\'{a}}n(2020)]{El-Kishky2020}
A.~El-Kishky and F.~Guzm{\'{a}}n.
\newblock {Massively Multilingual Document Alignment with Cross-lingual
  Sentence-Mover's Distance}.
\newblock 2020.
\newblock URL \url{http://arxiv.org/abs/2002.00761}.

\bibitem[El-Kishky and Guzm{\'a}n(2020)]{el2020massively}
A.~El-Kishky and F.~Guzm{\'a}n.
\newblock Massively multilingual document alignment with cross-lingual
  sentence-mover's distance.
\newblock \emph{arXiv preprint arXiv:2002.00761}, 2020.

\bibitem[{Farhath} et~al.(2018){Farhath}, {Ranathunga}, {Jayasena}, and
  {Dias}]{8421901}
F.~{Farhath}, S.~{Ranathunga}, S.~{Jayasena}, and G.~{Dias}.
\newblock Integration of bilingual lists for domain-specific statistical
  machine translation for sinhala-tamil.
\newblock In \emph{2018 Moratuwa Engineering Research Conference (MERCon)},
  pages 538--543, 2018.

\bibitem[Gale and Church(1993)]{gale-church-1993-program}
W.~A. Gale and K.~W. Church.
\newblock A program for aligning sentences in bilingual corpora.
\newblock \emph{Computational Linguistics}, 19\penalty0 (1):\penalty0 75--102,
  1993.
\newblock URL \url{https://www.aclweb.org/anthology/J93-1004}.

\bibitem[Gomes and {Pereira Lopes}(2016)]{Gomes2016}
L.~Gomes and G.~{Pereira Lopes}.
\newblock {First Steps Towards Coverage-Based Document Alignment}.
\newblock \emph{Proceedings of the First Conference on Machine Translation},
  2:\penalty0 697--702, 2016.
\newblock URL \url{http://www.aclweb.org/anthology/W/W16/W16-2369}.

\bibitem[Gr{\'{e}}goire and Langlais(2017)]{Gregoire2017}
F.~Gr{\'{e}}goire and P.~Langlais.
\newblock {BUCC 2017 Shared Task: a First Attempt Toward a Deep Learning
  Framework for Identifying Parallel Sentences in Comparable Corpora}.
\newblock In \emph{Proceedings of the 10th Workshop on Building and Using
  Comparable Corpora}, pages 46--50, Stroudsburg, PA, USA, 2017. Association
  for Computational Linguistics.
\newblock \doi{10.18653/v1/W17-2509}.
\newblock URL \url{http://aclweb.org/anthology/W17-2509}.

\bibitem[Guo et~al.(2019)Guo, Yang, Stevens, Cer, Ge, Sung, Strope, and
  Kurzweil]{Guo2019}
M.~Guo, Y.~Yang, K.~Stevens, D.~Cer, H.~Ge, Y.-h. Sung, B.~Strope, and
  R.~Kurzweil.
\newblock {Hierarchical Document Encoder for Parallel Corpus Mining}.
\newblock 1:\penalty0 64--72, 2019.
\newblock \doi{10.18653/v1/w19-5207}.

\bibitem[Guzm{\'a}n et~al.(2019)Guzm{\'a}n, Chen, Ott, Pino, Lample, Koehn,
  Chaudhary, and Ranzato]{guzman-etal-2019-flores}
F.~Guzm{\'a}n, P.-J. Chen, M.~Ott, J.~Pino, G.~Lample, P.~Koehn, V.~Chaudhary,
  and M.~Ranzato.
\newblock The {FLORES} evaluation datasets for low-resource machine
  translation: {N}epali{--}{E}nglish and {S}inhala{--}{E}nglish.
\newblock In \emph{Proceedings of the 2019 Conference on Empirical Methods in
  Natural Language Processing and the 9th International Joint Conference on
  Natural Language Processing (EMNLP-IJCNLP)}, pages 6098--6111, Hong Kong,
  China, Nov. 2019. Association for Computational Linguistics.
\newblock \doi{10.18653/v1/D19-1632}.
\newblock URL \url{https://www.aclweb.org/anthology/D19-1632}.

\bibitem[Joshi et~al.(2020)Joshi, Santy, Budhiraja, Bali, and
  Choudhury]{joshi2020state}
P.~Joshi, S.~Santy, A.~Budhiraja, K.~Bali, and M.~Choudhury.
\newblock The state and fate of linguistic diversity and inclusion in the nlp
  world.
\newblock \emph{arXiv preprint arXiv:2004.09095}, 2020.

\bibitem[Koehn and Knowles(2017)]{koehn2017six}
P.~Koehn and R.~Knowles.
\newblock Six challenges for neural machine translation.
\newblock \emph{arXiv preprint arXiv:1706.03872}, 2017.

\bibitem[Koehn et~al.(2019)Koehn, Guzman, Chaudhary, and Pino]{inproceedings}
P.~Koehn, F.~Guzman, V.~Chaudhary, and J.~Pino.
\newblock Findings of the wmt 2019 shared task on parallel corpus filtering for
  low-resource conditions.
\newblock pages 54--72, 01 2019.
\newblock \doi{10.18653/v1/W19-5404}.

\bibitem[Ma(2006)]{ma-2006-champollion}
X.~Ma.
\newblock {C}hampollion: A robust parallel text sentence aligner.
\newblock In \emph{Proceedings of the Fifth International Conference on
  Language Resources and Evaluation ({LREC}{'}06)}, Genoa, Italy, May 2006.
  European Language Resources Association (ELRA).

\bibitem[Mahata et~al.(2016)Mahata, Das, and Pal]{Mahata2016}
S.~Mahata, D.~Das, and S.~Pal.
\newblock {WMT2016: A Hybrid Approach to Bilingual Document Alignment}.
\newblock In \emph{Proceedings of the First Conference on Machine Translation:
  Volume 2, Shared Task Papers}, volume~2, pages 724--727, Stroudsburg, PA,
  USA, 2016. Association for Computational Linguistics.
\newblock \doi{10.18653/v1/W16-2373}.
\newblock URL \url{http://aclweb.org/anthology/W16-2373}.

\bibitem[Moore(2002)]{moore2002fast}
R.~C. Moore.
\newblock Fast and accurate sentence alignment of bilingual corpora.
\newblock In \emph{Conference of the Association for Machine Translation in the
  Americas}, pages 135--144. Springer, 2002.

\bibitem[Ott et~al.(2019)Ott, Edunov, Baevski, Fan, Gross, Ng, Grangier, and
  Auli]{ott-etal-2019-fairseq}
M.~Ott, S.~Edunov, A.~Baevski, A.~Fan, S.~Gross, N.~Ng, D.~Grangier, and
  M.~Auli.
\newblock fairseq: A fast, extensible toolkit for sequence modeling.
\newblock In \emph{Proceedings of the 2019 Conference of the North {A}merican
  Chapter of the Association for Computational Linguistics (Demonstrations)},
  pages 48--53, Minneapolis, Minnesota, June 2019. Association for
  Computational Linguistics.
\newblock \doi{10.18653/v1/N19-4009}.
\newblock URL \url{https://www.aclweb.org/anthology/N19-4009}.

\bibitem[Resnik(1998)]{resnik1998parallel}
P.~Resnik.
\newblock Parallel strands: A preliminary investigation into mining the web for
  bilingual text.
\newblock In \emph{Conference of the Association for Machine Translation in the
  Americas}, pages 72--82. Springer, 1998.

\bibitem[RESNIK(1999)]{resnik1999mining}
P.~RESNIK.
\newblock Mining the web for bilingual texts.
\newblock In \emph{Proceedings of the 37th Annual Meeting of the Association
  for Computational Linguistics, 1999}, 1999.

\bibitem[Resnik and Smith(2003)]{resnik2003web}
P.~Resnik and N.~A. Smith.
\newblock The web as a parallel corpus.
\newblock \emph{Computational Linguistics}, 29\penalty0 (3):\penalty0 349--380,
  2003.

\bibitem[Ruder et~al.(2019)Ruder, Vuli{\'c}, and S{\o}gaard]{ruder2019survey}
S.~Ruder, I.~Vuli{\'c}, and A.~S{\o}gaard.
\newblock A survey of cross-lingual word embedding models.
\newblock \emph{Journal of Artificial Intelligence Research}, 65:\penalty0
  569--631, 2019.

\bibitem[Schwenk and Douze(2017)]{schwenk2017learning}
H.~Schwenk and M.~Douze.
\newblock Learning joint multilingual sentence representations with neural
  machine translation.
\newblock \emph{arXiv preprint arXiv:1704.04154}, 2017.

\bibitem[Schwenk et~al.(2019)Schwenk, Chaudhary, Sun, Gong, and
  Guzm{\'{a}}n]{DBLP:journals/corr/abs-1907-05791}
H.~Schwenk, V.~Chaudhary, S.~Sun, H.~Gong, and F.~Guzm{\'{a}}n.
\newblock Wikimatrix: Mining 135m parallel sentences in 1620 language pairs
  from wikipedia.
\newblock \emph{CoRR}, abs/1907.05791, 2019.
\newblock URL \url{http://arxiv.org/abs/1907.05791}.

\bibitem[Thompson and Koehn(2019)]{thompson-koehn-2019-vecalign}
B.~Thompson and P.~Koehn.
\newblock {V}ecalign: Improved sentence alignment in linear time and space.
\newblock In \emph{Proceedings of the 2019 Conference on Empirical Methods in
  Natural Language Processing and the 9th International Joint Conference on
  Natural Language Processing (EMNLP-IJCNLP)}, pages 1342--1348, Hong Kong,
  China, Nov. 2019. Association for Computational Linguistics.
\newblock \doi{10.18653/v1/D19-1136}.
\newblock URL \url{https://www.aclweb.org/anthology/D19-1136}.

\bibitem[Uszkoreit et~al.(2010)Uszkoreit, Ponte, Popat, and
  Dubiner]{uszkoreit2010large}
J.~Uszkoreit, J.~M. Ponte, A.~C. Popat, and M.~Dubiner.
\newblock Large scale parallel document mining for machine translation.
\newblock In \emph{Proceedings of the 23rd International Conference on
  Computational Linguistics}, pages 1101--1109. Association for Computational
  Linguistics, 2010.

\bibitem[Varga et~al.(2007)Varga, Hal{\'a}csy, Kornai, Nagy, N{\'e}meth, and
  Tr{\'o}n]{varga2007parallel}
D.~Varga, P.~Hal{\'a}csy, A.~Kornai, V.~Nagy, L.~N{\'e}meth, and V.~Tr{\'o}n.
\newblock Parallel corpora for medium density languages.
\newblock \emph{Amsterdam Studies In The Theory And History Of Linguistic
  Science Series 4}, 292:\penalty0 247, 2007.

\bibitem[Volk et~al.(2010)Volk, Bubenhofer, Althaus, Bangerter, Furrer, and
  Ruef]{volkchallenges}
M.~Volk, N.~Bubenhofer, A.~Althaus, M.~Bangerter, L.~Furrer, and B.~Ruef.
\newblock Challenges in building a multilingual alpine heritage corpus.
\newblock In \emph{Proceedings of the Seventh International Conference on
  Language Resources and Evaluation ((LREC)'10)}, Valletta, Malta, May 2010.
  European Language Resources Association (ELRA).

\bibitem[Yeong et~al.(2019)Yeong, Tan, and Gan]{Yeong2019}
Y.-l. Yeong, T.-p. Tan, and K.~H. Gan.
\newblock {A Hybrid of Sentence-Level Approach and Fragment-Level Approach of
  Parallel Text Extraction from Comparable Text}.
\newblock \emph{Procedia Computer Science}, 161:\penalty0 406--414, 2019.
\newblock ISSN 1877-0509.
\newblock \doi{10.1016/j.procs.2019.11.139}.
\newblock URL \url{https://doi.org/10.1016/j.procs.2019.11.139}.

\bibitem[Zweigenbaum et~al.(2017)Zweigenbaum, Sharoff, and
  Rapp]{zweigenbaum-etal-2017-overview}
P.~Zweigenbaum, S.~Sharoff, and R.~Rapp.
\newblock Overview of the second {BUCC} shared task: Spotting parallel
  sentences in comparable corpora.
\newblock In \emph{Proceedings of the 10th Workshop on Building and Using
  Comparable Corpora}, pages 60--67, Vancouver, Canada, Aug. 2017. Association
  for Computational Linguistics.
\newblock \doi{10.18653/v1/W17-2512}.
\newblock URL \url{https://www.aclweb.org/anthology/W17-2512}.

\end{thebibliography}

\end{document}